\documentclass{llncs}
\pdfoutput=1
\usepackage{makeidx}

\usepackage{amsmath}
\usepackage{amsfonts}
\usepackage{amssymb}

\usepackage{psfrag}
\usepackage{cite}
\usepackage{color}
\usepackage{framed}
\definecolor{shadecolor}{gray}{0.9}
\usepackage[ruled]{algorithm2e}
\usepackage{graphicx}
\newcommand{\comment}[1]{}
\begin{document}

\title{A Linear Time Natural Evolution Strategy for Non-Separable Functions}
\author{Yi Sun \and Faustino Gomez \and Tom Schaul \and J\"urgen Schmidhuber}
\institute{IDSIA, University of Lugano \& SUPSI, Galleria 2, Manno, CH-6928, Switzerland \\ \email{\{yi,tino,tom,juergen\}@idsia.ch}}
\maketitle 

\begin{abstract}
We present a novel Natural Evolution Strategy (NES) variant,
the Rank-One NES (R1-NES), which uses a 
low rank approximation of the search distribution covariance matrix. The algorithm
allows computation of the natural gradient with cost linear in the dimensionality of
the parameter space, and excels in solving high-dimensional
non-separable problems, including the best result to date on the
Rosenbrock function (512 dimensions).
\end{abstract}



\newcommand{\assign}{$\leftarrow$}
\section{Introduction}

Black-box optimization (also called zero-order optimization) methods have
received a great deal of attention in recent years due to their broad
applicability to real world
problems~\cite{health,quantum,chromatography,control,aeronautic,nozzle}. When
the structure of the objective function is unknown or too complex to model
directly, or when gradient information is unavailable or unreliable, such
methods are often seen as the last resort because all they require is that the
objective function can be evaluated at specific points.

In continuous black-box optimization problems, the state-of-the-art
algorithms, such as xNES~\cite{10-GECCOXNES} and CMA-ES~\cite{hansen:2001},
are all based the same principle~\cite{10-GECCOXNES,Akimoto2010ppsn}: a
Gaussian search distribution is repeatedly updated based on the objective
function values of sampled points. Usually the full covariance matrix of the
distribution is updated, allowing the algorithm to adapt the size and shape of
the Gaussian to the local characteristics of the objective function. Full
parameterization also provides invariance under affine transformations of the
coordinate system,
so that ill-shaped, highly non-separable problems can be tackled. However,
this generality comes with a price. The number of parameters scales
quadratically in the number of dimensions, and the computational cost per
sample is at least quadratic in the number of dimensions \cite{ros:2008}, and
sometimes cubic~\cite{10-GECCOXNES,09-ICMLENES}.

This cost is often justified since evaluating the objective function can
dominate the computation, and thus the main focus is on the improvement of
sampling efficiency. However, there are many problems, such as optimizing
weights in neural networks where the dimensionality of the parameter space can
be very large (e.g.\ many thousands of weights), the quadratic cost of
updating the search distribution can become the computational bottleneck. One
possible remedy is to restrict the covariance matrix to be
diagonal~\cite{Schaul2011snes}, which reduces the computation per function
evaluation to $O(d)$, linear in the number $d$ of dimensions. Unfortunately,
this ``diagonal'' approach performs poorly when the problem is non-separable
because the search distribution cannot follow directions that are not parallel
to the current coordinate axes. \comment{ Other remedies include using
block diagonal covariance matrices, however, identifying sub-blocks
usually requires specific knowledge about the problem structure
\{cite tino\}.  }

In this paper, we propose a new variant of the natural evolution strategy
family~\cite{wierstra:2008}, termed Rank One NES (R1-NES). This algorithm
stays within the general NES framework in that the search distribution is
adjusted according to the natural gradient \cite{amari98natural}, but it uses
a novel parameterization of the covariance matrix,%
\[
C=\sigma^{2}\left(  I+uu^{\top}\right)  \text{,}%
\]
where $u$ and $\sigma$ are the parameters to be adjusted. This
parameterization allows the predominant eigen direction, $u$, of $C$ to be
aligned in any direction, enabling the algorithm to tackle highly
non-separable problems while maintaining only $O\left(  d\right)  $
parameters. We show through rigorous derivation that the natural gradient can
also be effectively computed in $O\left(  d\right)  $ per sample. R1-NES
scales well to high dimensions, and dramatically outperforms diagonal
covariance matrix algorithms on non-separable objective functions. As an
example, R1-NES reliably solves the the non-convex Rosenbrock function up to
$512$ dimensions.

The rest of the paper is organized as follows. Sectio~\ref{sec:nes}, briefly
reviews the NES framework. The derivation of R1-NES is presented in
Section~\ref{sec:r1-nes}. Section~\ref{sec:exp} empirically evaluates the
algorithm on standard benchmark functions, and Section~\ref{sec:conclusion}
concludes the paper.

\section{The NES framework}

\label{sec:nes} Natural evolution strategies (NES) are a class of evolutionary
algorithms for real-valued optimization that maintain a search distribution,
and adapt the distribution parameters by following the \emph{natural} gradient
of the expected function value. The success of such algorithms is largely
attributed to the use of natural gradient, which has the advantage of always
pointing in the direction of the steepest ascent, even if the parameter space
is not Euclidean. Moreover, compared to regular gradient, natural gradient
reduces the weights of gradient components with higher uncertainty, therefore
making the update more reliable. As a consequence, NES algorithms can
effectively cope with objective functions with ill-shaped landscapes,
especially preventing premature convergence on plateaus and avoiding
overaggressive steps on ridges~\cite{09-ICMLENES}.

\begin{figure}[t]
\centerline{
\psfrag{mu}[l][l]{$\mu$}
\psfrag{sigma}[c][c]{$\sigma$}
\includegraphics[width=\textwidth]{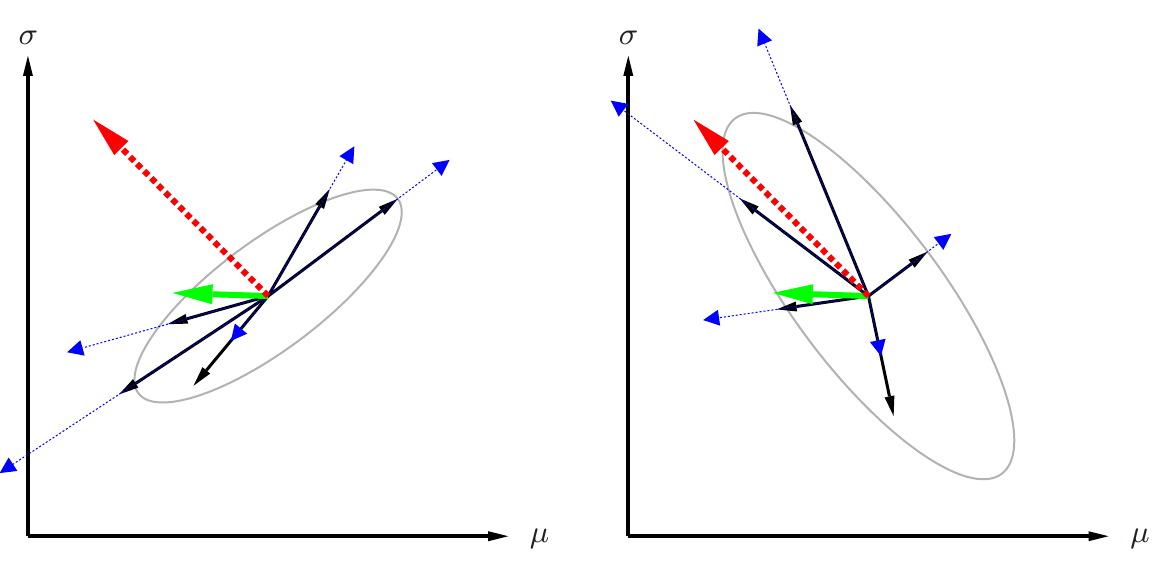}
} \caption[Illustration of plain vs. natural gradient]{\textbf{Plain versus
natural gradient in parameter space.} Consider two parameters, e.g.
$\theta=(\mu,\sigma)$, of the search distribution. On the left, the solid
(black) arrows indicate the gradient samples $\nabla_{\theta}\log
\pi(x\,|\,\theta)$, the dotted (blue) arrows correspond to $f(x)\cdot
\nabla_{\theta}\log\pi(x\,|\,\theta)$, that is, the same gradient estimates,
but scaled with fitness. Combining these, the bold (green) arrow indicates the
(sampled) fitness gradient $\nabla_{\theta}J$, while the bold dashed (red)
arrow indicates the corresponding natural gradient $\tilde{\nabla}_{\theta
}J=F^{-}\triangledown_{\theta}J$. Being random variables with expectation
zero, the distribution of the black arrows is governed by their covariance
(the gray ellipse). Note that this covariance is a quantity in \emph{parameter
space} (where the $\theta$ reside); not to be confused with that of the
\emph{search space} (where the samples $x$ reside). In contrast, on the right,
the solid (black) arrows represent $\tilde{\nabla}_{\theta}\log\pi
(x\,|\,\theta)$, and dotted (blue) arrows indicate the \emph{natural} gradient
samples $f(x)\cdot\tilde{\nabla}_{\theta}\log\pi(x\,|\,\theta)$, resulting in
the natural gradient (dashed red). The covariance of the solid arrows on the
right hand side turns out to be the inverse of the covariance of the solid
arrows on the left. This has the effect that when computing the natural
gradient, directions with high variance (uncertainty) are penalized and thus
shrunken, while components with low variance (high certainty) are boosted,
since these components of the gradient samples deserve more trust. This makes
the (dashed red) natural gradient a much more trustworthy update direction
than the (green) plain gradient. }%
\label{natgrad-illust}%
\end{figure}

The general framework of NES is given as follows: At each time step, the
algorithm samples $n$ $\in\mathbb{N}$ new samples $x_{1},\dots,x_{n}\sim
\pi\left(  \cdot|\theta\right)  $, with $\pi\left(  \cdot|\theta\right)  $
being the search distribution parameterized by $\theta$. Let $f:\mathbb{R}%
^{d}\mapsto\mathbb{R}$ be the objective function to maximize. The expected
function value under the search distribution is%
\[
J\left(  \theta\right)  =\mathbb{E}_{\theta}\left[  f\left(  x\right)
\right]  =\int f\left(  x\right)  \pi\left(  x|\theta\right)  dx\text{.}%
\]
Using the log-likelihood trick, the gradient w.r.t. the parameters can be
written as%
\begin{align*}
\triangledown_{\theta}J  &  =\triangledown_{\theta}\int f\left(  x\right)
\pi\left(  x|\theta\right)  dx\\
&  =\int f\left(  x\right)  \pi\left(  x|\theta\right)  \frac{\triangledown
_{\theta}\pi\left(  x|\theta\right)  }{\pi\left(  x|\theta\right)  }dx\\
&  =E\left[  f\left(  x\right)  \triangledown_{\theta}\log\pi\left(
x|\theta\right)  \right]  \text{,}%
\end{align*}
from which we obtain the Monte-Carlo estimate%
\[
\triangledown_{\theta}J\simeq\frac{1}{n}\sum_{i=1}^{n}f\left(  x_{i}\right)
\triangledown_{\theta}\log\pi\left(  x_{i}|\theta\right)  \text{.}%
\]
of the search gradient. The key step of NES then consists of replacing this
gradient by the natural gradient%
\[
\tilde{\triangledown}_{\theta}J=F^{-}\triangledown_{\theta}J\text{,}%
\]
where
\[
F=E\left[  \triangledown_{\theta}\log\pi\left(  x_{i}|\theta\right)
\triangledown_{\theta}\log\pi\left(  x_{i}|\theta\right)  ^{\top}\right]
\]
is the Fisher information matrix (See Fig.\ref{natgrad-illust} for an
illustration). This leads to a straightforward scheme of natural gradient
ascent for iteratively updating the parameters%
\begin{align}
\theta &  \leftarrow\theta+\eta\tilde{\triangledown}_{\theta}J\nonumber\\
&  =\theta+\frac{\eta}{n}\sum_{i=1}^{n}f\left(  x_{i}\right)  F^{-}%
\triangledown_{\theta}\log\pi\left(  x_{i}|\theta\right) \nonumber\\
&  =\theta+\frac{\eta}{n}\sum_{i=1}^{n}f\left(  x_{i}\right)  \tilde
{\triangledown}_{\theta}\log\pi\left(  x_{i}|\theta\right)  \text{.}
\label{eqnat}%
\end{align}
The sequence of 1) sampling an offspring population, 2) computing the
corresponding Monte Carlo estimate of the gradient, 3) transforming it into
the natural gradient, and 4) updating the search distribution, constitutes one
iteration of NES.

The most difficult step in NES is the computation of the Fisher information
matrix with respect to the parameterization. For full Gaussian distribution,
the Fisher can be derived analytically \cite{10-GECCOXNES,09-ICMLENES}.
However, for arbitrary parameterization of $C$, the Fisher matrix can be
highly non-trivial.


\section{Natural gradient of the rank-one covariance matrix approximation}

\label{sec:r1-nes}

In this paper, we consider a special formulation of the covariance matrix%
\[
C=\sigma^{2}\left(  I+uu^{\top}\right)  \text{,}%
\]
with parameter set $\theta=\left\langle \sigma,u\right\rangle $. The special
part of the parameterization is the vector $u\in\mathbb{R}^{d}$, which
corresponds to the predominant direction of $C$. This allows the search
distribution to be aligned in any direction by adjusting $u$, enabling the
algorithm to follow valleys not aligned with the current coordinate axes,
which is essential for solving non-separable problems.

Since $\sigma$ should always be positive, following the same procedure in
\cite{10-GECCOXNES}, we parameterize $\sigma=e^{\lambda}$, so that $\lambda
\in\mathbb{R}$ can be adjusted freely using gradient descent without worrying
about $\sigma$ becoming negative. The parameter set is adjusted to
$\theta=\left\langle \lambda,u\right\rangle $ accordingly.

From the derivation of \cite{09-ICMLENES}, the natural gradient on the sample
mean is given by%
\begin{equation}
\tilde{\triangledown}_{\mu}\log p\left(  x|\theta\right)  =x-\mu\text{.}
\label{grdnat_mu}%
\end{equation}
In the subsequent discussion we always assume $\mu=0$ for simplicity. It is
straightforward to sample from $\mathcal{N}\left(  0,C\right)  $\footnote{For
succinctness, we always assume the mean of the search distribution is $0$.
This can be achieved easily by shifting the coordinates.} by lettting
$y\sim\mathcal{N}\left(  0,I\right)  $, $z\sim\mathcal{N}\left(  0,1\right)
$, then%
\[
x=\sigma\left(  y+zu\right)  \sim\mathcal{N}\left(  0,C\right)  \text{.}%
\]
The inverse of $C$ can also be computed easily as%
\[
C^{-}=\sigma^{-2}\left(  I-\frac{1}{1+r^{2}}uu^{\top}\right)  \text{,}%
\]
where $r^{2}=u^{\top}u$. Using the relation $\det\left(  I+uu^{\top}\right)
=1+u^{\top}u$, the determinant of $C$ is%
\[
\left\vert C\right\vert =\sigma^{2d}\left(  1+r^{2}\right)  \text{.}%
\]

Knowing $C^{-}$ and $\left\vert C\right\vert $ allows the log-likelihood to be
written explicitly as%
\begin{align*}
\log p\left(  x|\theta\right)   &  =const-\frac{1}{2}\log\left\vert
C\right\vert -\frac{1}{2}x^{\top}C^{-}x\\
&  =const-\lambda d-\frac{1}{2}\log\left(  1+r^{2}\right)  -\frac{1}%
{2}e^{-2\lambda}x^{\top}x+\frac{1}{2}\frac{e^{-2\lambda}}{1+r^{2}}\left(
x^{\top}u\right)  ^{2}\text{.}%
\end{align*}
The regular gradient with respect to $\lambda$ and $u$ can then be computed
as:%
\begin{align}
\triangledown_{\lambda}\log p\left(  x|\theta\right)   &  =-d+e^{-2\lambda
}\left(  x^{\top}x-\frac{\left(  x^{\top}u\right)  ^{2}}{1+r^{2}}\right)
\text{,}\label{grd_lambda}\\
\triangledown_{u}\log p\left(  x|\theta\right)   &  =-\frac{u}{1+r^{2}%
}+e^{-2\lambda}\left[  -\frac{\left(  x^{\top}u\right)  ^{2}u}{\left(
1+r^{2}\right)  ^{2}}+\frac{\left(  x^{\top}u\right)  x}{1+r^{2}}\right]
\text{.} \label{grd_u}%
\end{align}

Replacing $x$ with $e^{\lambda}\left(  y+zu\right)  $, then the Fisher can be
computed by marginalizing out i.i.d. standard Gaussian variables $y$ and $z$,
namely,%
\begin{align*}
F  &  =E_{x}\left[  \triangledown_{\theta}\log p\left(  x|\theta\right)
\triangledown_{\theta}\log p\left(  x|\theta\right)  ^{\top}\right] \\
&  =E_{y,z}\left[  \triangledown_{\theta}\log p\left(  y+zu|\theta\right)
\triangledown_{\theta}\log p\left(  y+zu|\theta\right)  ^{\top}\right]
\text{.}%
\end{align*}
Since elements in $\triangledown_{\theta}\log p\left(  x|\theta\right)
\triangledown_{\theta}\log p\left(  x|\theta\right)  ^{\top}$ are essentially
polynomials of $y$ and $z$, their expectations can be computed
analytically\footnote{The derivation is tedious, thus omitted here. All
derivations are numerically verified using Monte-Carlo simulation.}, which
gives the exact Fisher information matrix%
\[
F=\left[
\begin{array}
[c]{cc}%
2d & \frac{2u^{\top}}{1+r^{2}}\\
\frac{2u}{1+r^{2}} & B
\end{array}
\right]  \text{,}%
\]
with%
\[
B=\frac{1}{1+r^{2}}\left[  r^{2}I+\frac{1-r^{2}}{1+r^{2}}uu^{\top}\right]
\text{.}%
\]

Let $v=u/r$, then%
\[
F=\frac{r^{2}}{1+r^{2}}\left[
\begin{array}
[c]{cc}%
2d\frac{1+r^{2}}{r^{2}} & 2\frac{v^{\top}}{r}\\
2\frac{v}{r} & I+\frac{1-r^{2}}{1+r^{2}}vv^{\top}%
\end{array}
\right]  \text{.}%
\]
The inverse of $F$ is thus given by%
\[
F^{-}=\frac{1+r^{2}}{r^{2}}\left[
\begin{array}
[c]{cc}%
2d\frac{1+r^{2}}{r^{2}} & 2\frac{v^{\top}}{r}\\
2\frac{v}{r} & I+\frac{1-r^{2}}{1+r^{2}}vv^{\top}%
\end{array}
\right]  ^{-}\text{.}%
\]
We apply the formula for block matrix inverse in \cite{08-MatCookbook}%
\[
\left[
\begin{array}
[c]{cc}%
A_{11} & A_{12}\\
A_{21} & A_{22}%
\end{array}
\right]  ^{-}=\left[
\begin{array}
[c]{cc}%
C_{1}^{-} & -A_{11}^{-}A_{12}C_{2}^{-}\\
-C_{2}^{-}A_{21}A_{11}^{-} & C_{2}^{-}%
\end{array}
\right]  \text{,}%
\]
where $C_{1}=A_{11}-A_{12}A_{22}^{-}A_{21}$, and $C_{2}=A_{22}-A_{21}%
A_{11}^{-}A_{12}$ are the Schur complements. Let $F$ be partitioned as above,
then%
\[
B^{-}=I-\frac{1-r^{2}}{2}vv^{\top}\text{,}%
\]
and the Shur complements are%
\begin{align*}
C_{1}  &  =2d\frac{1+r^{2}}{r^{2}}-4\frac{v^{\top}}{r}\left(  I-\frac{1-r^{2}%
}{2}vv^{\top}\right)  \frac{v}{r}\\
&  =2d\left(  \frac{1+r^{2}}{r^{2}}\right)  -2\frac{1+r^{2}}{r^{2}}\\
&  =2\left(  d-1\right)  \left(  \frac{1+r^{2}}{r^{2}}\right)  \text{,}%
\end{align*}
and%
\begin{align*}
C_{2}  &  =I+\frac{1-r^{2}}{1+r^{2}}vv^{\top}-\frac{2vv^{\top}}{d\left(
1+r^{2}\right)  }\\
&  =I+\frac{1}{1+r^{2}}\left[  1-r^{2}-\frac{2}{d}\right]  vv^{\top}\text{,}%
\end{align*}
whose inverse is given by%
\[
C_{2}^{-}=I+\frac{2+d\left(  r^{2}-1\right)  }{2\left(  d-1\right)  }vv^{\top
}\text{.}%
\]
Combining the results gives the analytical solution of the inverse Fisher:%
\[
F^{-}=\frac{1+r^{2}}{2r^{2}\left(  d-1\right)  }\left[
\begin{array}
[c]{cc}%
\frac{r^{2}}{1+r^{2}} & \ \ -rv^{\top}\\
-rv & 2\left(  d-1\right)  \ \ I+\left[  2+d\left(  r^{2}-1\right)  \right]
vv^{\top}%
\end{array}
\right]  \text{.}%
\]
Multiplying $F^{-}$ with the regular gradient in Eq.\ref{grd_lambda} and
Eq.\ref{grd_u} gives the natural gradient for $\lambda$ and $u$:%
\begin{equation}
\tilde{\triangledown}_{\lambda}\log p\left(  x|\theta\right)  =\frac
{1}{2\left(  d-1\right)  }\left[  \left(  e^{-2\lambda}x^{\top}x-d\right)
-\left(  e^{-2\lambda}\left(  x^{\top}v\right)  ^{2}-1\right)  \right]
\text{.} \label{grdnat_lambda}%
\end{equation}
and%
\begin{equation}
\tilde{\triangledown}_{u}\log p\left(  x|\theta\right)  =\frac{e^{-2\lambda}%
}{2\left(  d-1\right)  r}\left[  \left(  1-d\right)  \left(  x^{\top}v\right)
^{2}+\left(  r^{2}+1\right)  \left(  \left(  x^{\top}v\right)  ^{2}-x^{\top
}x\right)  \right]  \text{.} \label{grdnat_u}%
\end{equation}

Note that computing both $\tilde{\triangledown}_{\lambda}\log p\left(
x|\theta\right)  $ and $\tilde{\triangledown}_{u}\log p\left(  x|\theta
\right)  $ requires only the inner products $x^{\top}x$ and $x^{\top}v$,
therefore can be done $O\left(  d\right)  $ storage and time.

\SetAlgoNoLine
\SetKwComment{dbc}{//}{} \SetFuncSty{sc}
\LinesNumbered
\DontPrintSemicolon
\SetAlCapNameSty{sc} \SetProcNameSty{sc}

\begin{algorithm}
[t]\caption{ R1-NES($n$) } \label{alg:r1-nes} \While{not terminate}{
\For{i = 1 to n}{
$y_i$ \assign $\mathcal{N}(0,I)$ \;
$z_i$ \assign $\mathcal{N}(0,1)$\;
$x_i$ \assign $e^{\lambda}(y_i+z_iu)$ \dbc*[r]{generate sample}
fitness[i] \assign $f(\mu + x_i)$\;
}
Compute the natural gradient for $\mu$, $\lambda$, $u$, $c$, and $v$
according to Eq.\ref{grdnat_mu}, \ref{grdnat_lambda}, \ref{grdnat_u},
\ref{grdnat_c}, and \ref{grdnat_v}, and combine them using Eq.\ref{eqnat}\;
$\mu$ \assign $\mu+\eta\tilde{\triangledown}_{\mu}J$\;
$\lambda$ \assign
$\lambda+\eta\tilde{\triangledown}_{\lambda}J$\;
\eIf{$\tilde{\triangledown}_{c}\log p\left(  x|\theta\right)<0$}
{$c$  \assign  $c+\eta\tilde{\triangledown}_{c}J$\;
$v$  \assign  $\frac{v+\eta\tilde{\triangledown}_{v}J}{\left\Vert
v+\eta\tilde{\triangledown}_{v}J\right\Vert }$\;
$u$ \assign $e^{c}v$}
{
$u$ \assign $u+\eta\tilde{\triangledown}_{u}J$ \dbc*[r]{additive update}
$c$ \assign $\log\left\Vert u\right\Vert$\;
$v$ \assign $\frac{u}{\left\Vert u\right\Vert }$
}
}

\end{algorithm}


\subsection{Reparameterization}

\begin{figure}[t]
\begin{center}
\includegraphics[width=0.5\textwidth]{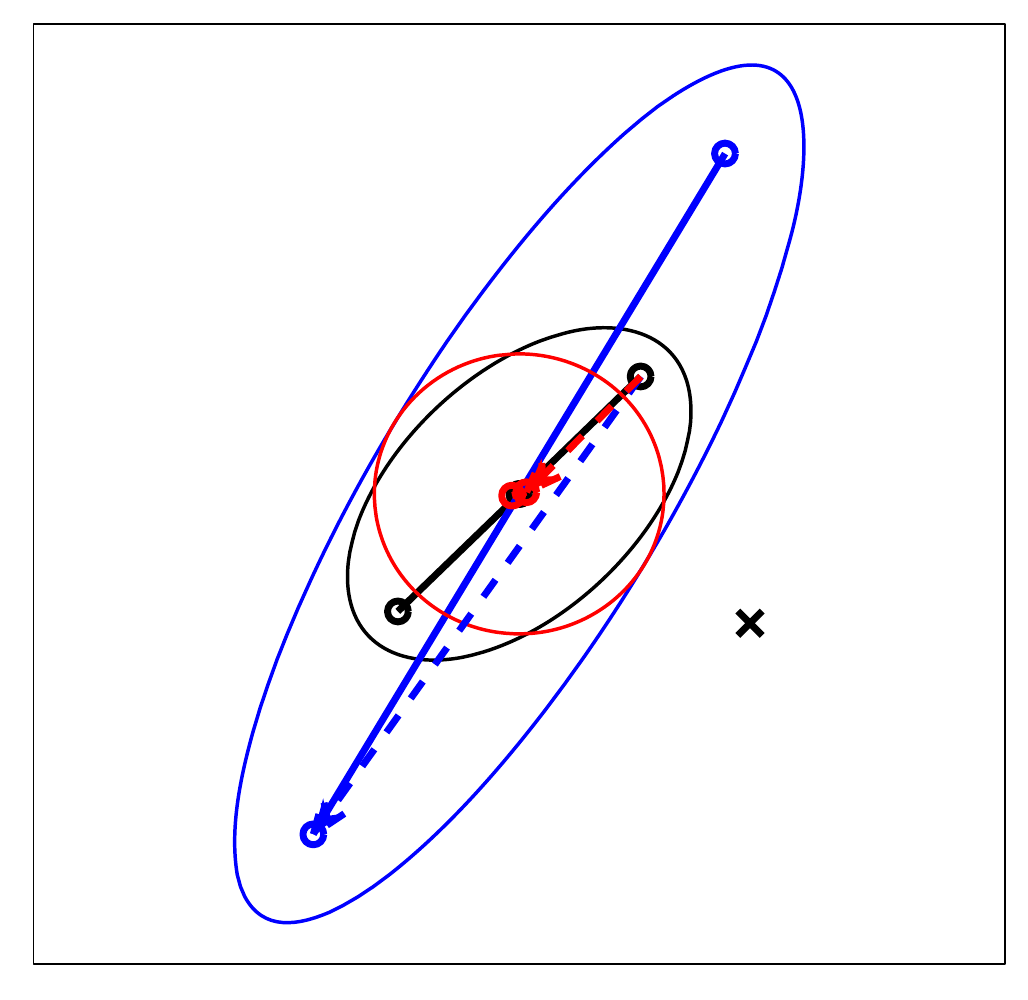}\hspace*{-2mm}
\includegraphics[width=0.5\textwidth]{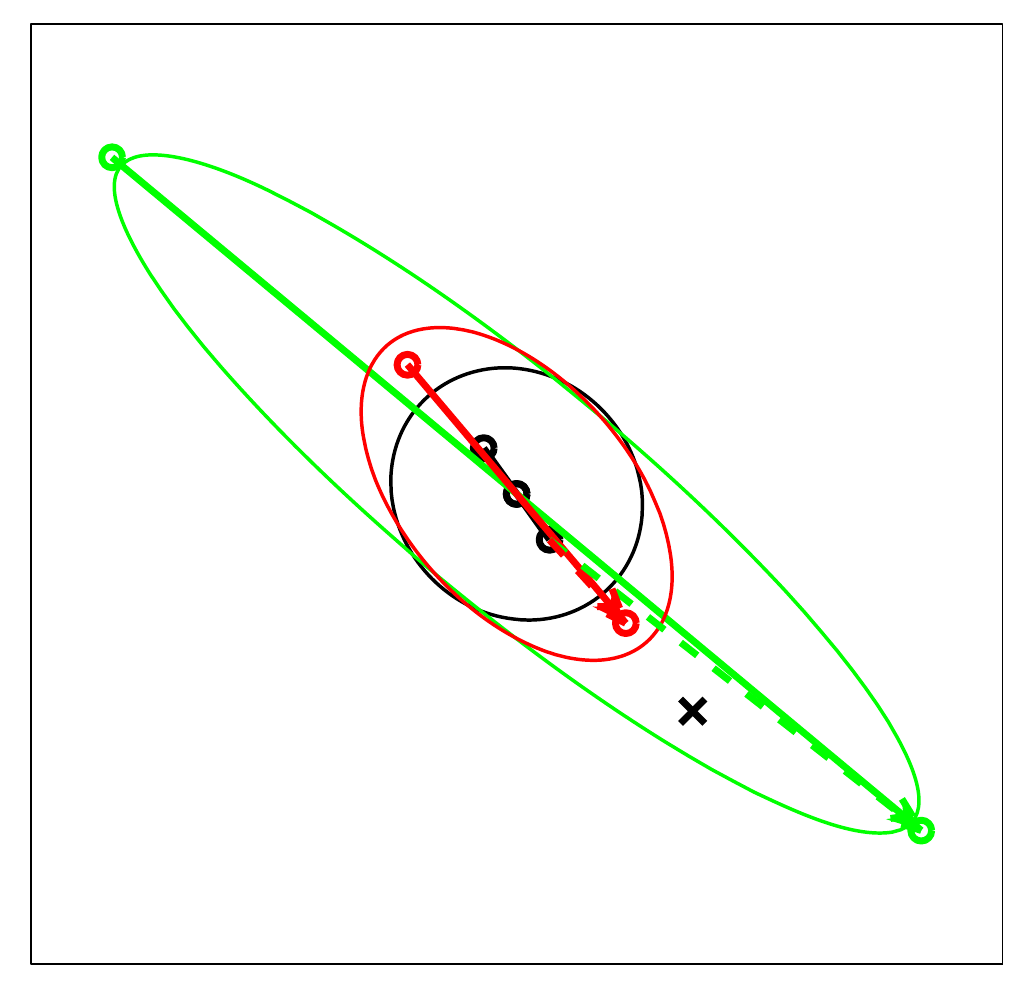}
\end{center}
\par
\vspace*{-4mm} \hspace*{1.1in} (a) \hspace*{2.15in} (b)
\caption{\textbf{Illustration of the change in parameterization.} In both
panels, the black lines and ellipses refer to the current predominant
direction $u$, and the corresponding search distribution from which samples
are drawn. The black cross denotes one such sample that is being used to
update the distribution. In the left panel, the direction of the selected
point is almost perpendicular to $u$, resulting in a large gradient, reducing
$u$ (the dotted blue line). However, direct gradient update on $u$ will flip
the direction of $u$. As a result, $u$ stays in the same undesired direction,
but with increased length. In contrast, performing update on $c$ and $v$ gives
the predominant search direction depicted in the red, with $u$ shrunk
properly. The right panel shows another case where the selected point aligns
with the search direction, and performing the exponential update on $c$ and
$v$ causes $u$ to increase dramatically (green line \& ellipsoid). This effect
is prevented by performing the additive update (Eq.~\ref{grdnat_u}) on $u$
(red line \& ellipsoid).}%
\label{fig:param}%
\end{figure}

The natural gradient above is obtained with respect to $u$. However, direct
gradient update on $u$ has an unpleasant property when $\tilde{\triangledown
}_{u}\log p\left(  x|\theta\right)  $ is in the opposite direction of $u$,
which is illustrated in Fig.~\ref{fig:param}(a). In this case, the gradient
tends to shrink $u$. However, if $\tilde{\triangledown}_{u}\log p\left(
x|\theta\right)  $ is large, adding the gradient will flip the direction of
$u$, and the length of $u$ might even grow. This causes numerical problems,
especially when $r$ is small. A remedy is to separate the length and direction
of $u$, namely, reparameterize $u=e^{c}v$, where $\left\Vert v\right\Vert =1$
and $e^{c}$ is the length of $u$. Then the gradient update on $c$ will never
flip $u$, and thus avoid the problem.

Note that for small change $\delta u$, the update on $c$ and $v$ can be
obtained from%
\begin{align*}
\delta c  &  =\frac{1}{2}\log\left(  u+\delta u\right)  ^{\top}\left(
u+\delta u\right)  -c\\
&  \simeq\frac{1}{2}\log\left(  u^{\top}u+2\delta u^{\top}u\right)  -c\\
&  =\frac{1}{2}\log u^{\top}u+\frac{1}{2}\log\left(  1+\frac{2\delta u^{\top
}u}{u^{\top}u}\right)  -c\\
&  \simeq\frac{\delta u^{\top}u}{u^{\top}u}%
\end{align*}
and%
\begin{align*}
\delta v  &  =\frac{u+\delta u}{\sqrt{\left(  u+\delta u\right)  ^{\top
}\left(  u+\delta u\right)  }}-v\\
&  \simeq\frac{u+\delta u}{\left(  u^{\top}u+2\delta u^{\top}u\right)
^{\frac{1}{2}}}-\frac{u}{\left(  u^{\top}u\right)  ^{\frac{1}{2}}}\\
&  =\frac{u+\delta u}{\left(  u^{\top}u\right)  ^{\frac{1}{2}}\left(
1+\frac{2\delta u^{\top}u}{u^{\top}u}\right)  ^{\frac{1}{2}}}-\frac{u}{\left(
u^{\top}u\right)  ^{\frac{1}{2}}}\\
&  \simeq\frac{\left(  u+\delta u\right)  \left(  1-\frac{\delta u^{\top}%
u}{u^{\top}u}\right)  }{\left(  u^{\top}u\right)  ^{\frac{1}{2}}}-\frac
{u}{\left(  u^{\top}u\right)  ^{\frac{1}{2}}}\\
&  =\frac{1}{\left(  u^{\top}u\right)  ^{\frac{1}{2}}}\left[  \delta
u-\frac{\delta u^{\top}u}{u^{\top}u}u\right]  \text{.}%
\end{align*}
The natural gradient on $c$ and $v$ is given by letting $\delta u\propto
\tilde{\triangledown}_{u}\log p\left(  x|\theta\right)  $, thanks to the
invariance property:%
\begin{align}
\tilde{\triangledown}_{c}\log p\left(  x|\theta\right)   &  =r^{-1}%
\tilde{\triangledown}_{u}\log p\left(  x|\theta\right)  ^{\top}%
v\label{grdnat_c}\\
\tilde{\triangledown}_{v}\log p\left(  x|\theta\right)   &  =r^{-1}\left[
\tilde{\triangledown}_{u}\log p\left(  x|\theta\right)  -\left(
\tilde{\triangledown}_{u}\log p\left(  x|\theta\right)  ^{\top}v\right)
v\right]  \text{,} \label{grdnat_v}%
\end{align}
Note that computing $\tilde{\triangledown}_{c}\log p\left(  x|\theta\right)  $
and $\tilde{\triangledown}_{v}\log p\left(  x|\theta\right)  $ involves only
inner products between vectors, which can also be done linearly in the number
of dimensions.

Using the parameterization $\left\langle c,v\right\rangle $ introduces another
problem. When $r$ is small, $\tilde{\triangledown}_{c}\log p\left(
x|\theta\right) $ tends to be large, and thus directly updating $c$ causes $r$
to grow exponentially, resulting in numerical instability, as shown in
Fig.~\ref{fig:param}(b). In this case, the additive update on $u$, rather than
the update on $\left\langle c,v\right\rangle $ is more stable. In our
implementation, the additive update on $u$ is used if $\tilde{\triangledown
}_{c}\log p\left(  x|\theta\right)  >0$, otherwise the update is on
$\left\langle c,v\right\rangle $.
This solution proved to be numerially stable in all our tests.
Algorithm~\ref{alg:r1-nes} shows the complete R1-NES algorithm in pseudocode.

\section{Experiments}

\label{sec:exp}

The R1-NES algorithm was evaluated on the twelve noise-free unimodal
functions~\cite{bbobnoisefree} in the `Black-Box Optimization Benchmarking'
collection (BBOB) from the 2010 GECCO Workshop for Real-Parameter
Optimization. In order to make the results comparable those of other methods,
the setup in~\cite{bbobsetup} was used, which transforms the pure benchmark
functions to make the parameters non-separable (for some) and avoid trivial
optima at the origin.

R1-NES was compared to xNES~\cite{Glasmachers2010a},
SNES~\cite{Schaul2011snes} on each benchmark with problem dimensions $d=2^{k}%
$, $k=\{1..9\}$ (20 runs for each setup), except for xNES, which was only run
up $k=6, d=64$. Note that xNES serves as a proper baseline since it is
state-of-the-art, achieving performance on par with the popular CMA-ES. The
reference machine is an Intel Core i7 processor with 1.6GHz and 4GB of RAM.

\begin{figure}[h]
\centerline{
		\includegraphics[width=1.1\textwidth]{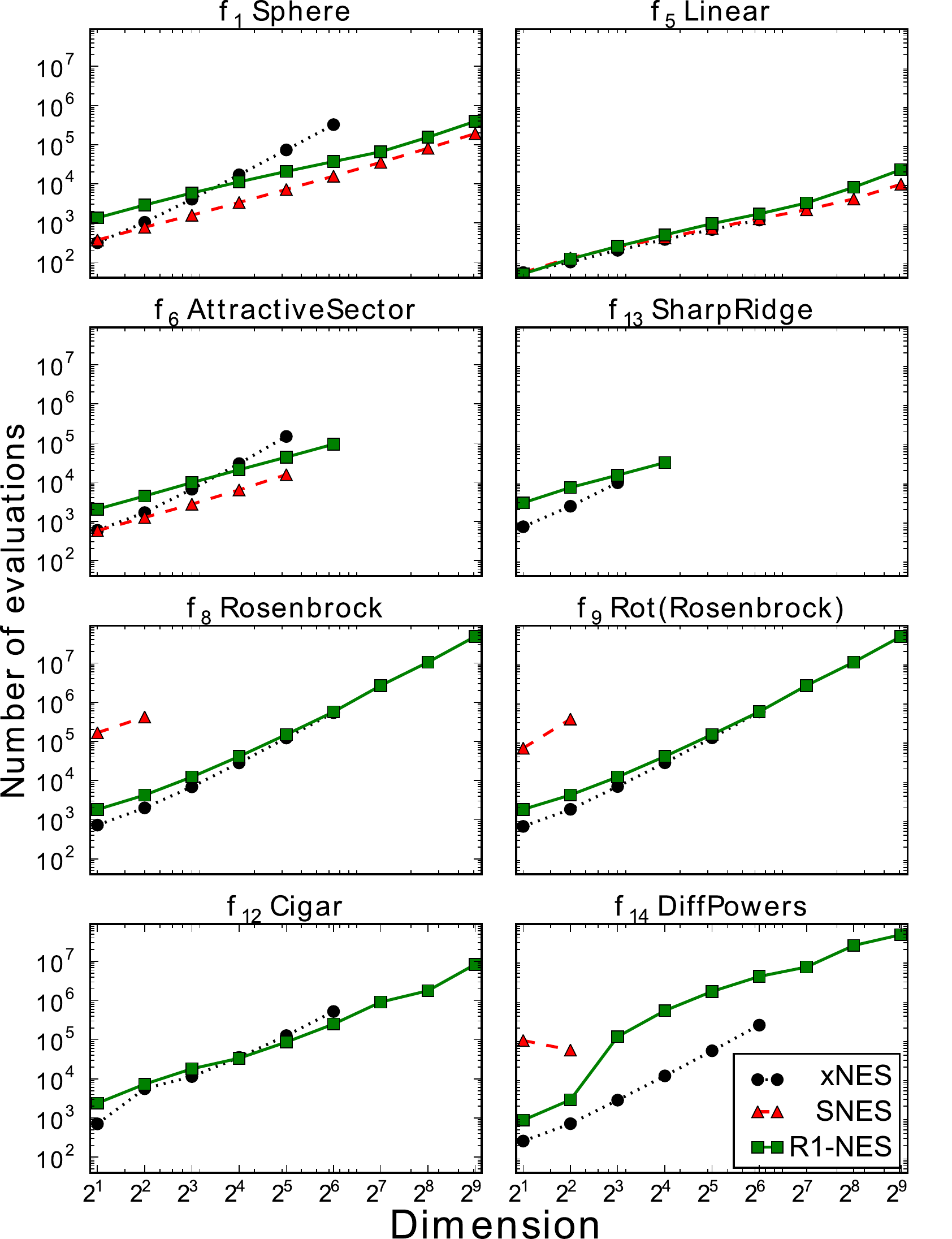}
	} \caption{\textbf{Performance comparison on BBOB unimodal benchmarks.}
Log-log plot of the median number of fitness evaluations (over 20 trials)
required to reach the target fitness value of $-10^{-8}$ for unimodal
benchmark functions for which R1-NES is well suited, on dimensions 2 to 512
(cases for which 90\% or more of the runs converged prematurely are not
shown). Note that xNES consistently solves all benchmarks on small dimensions
($\le64$), with a scaling factor that is almost the same over all functions. }%
\label{fig:good}%
\end{figure}

\begin{figure}[t]
\centerline{
		\includegraphics[width=\columnwidth]{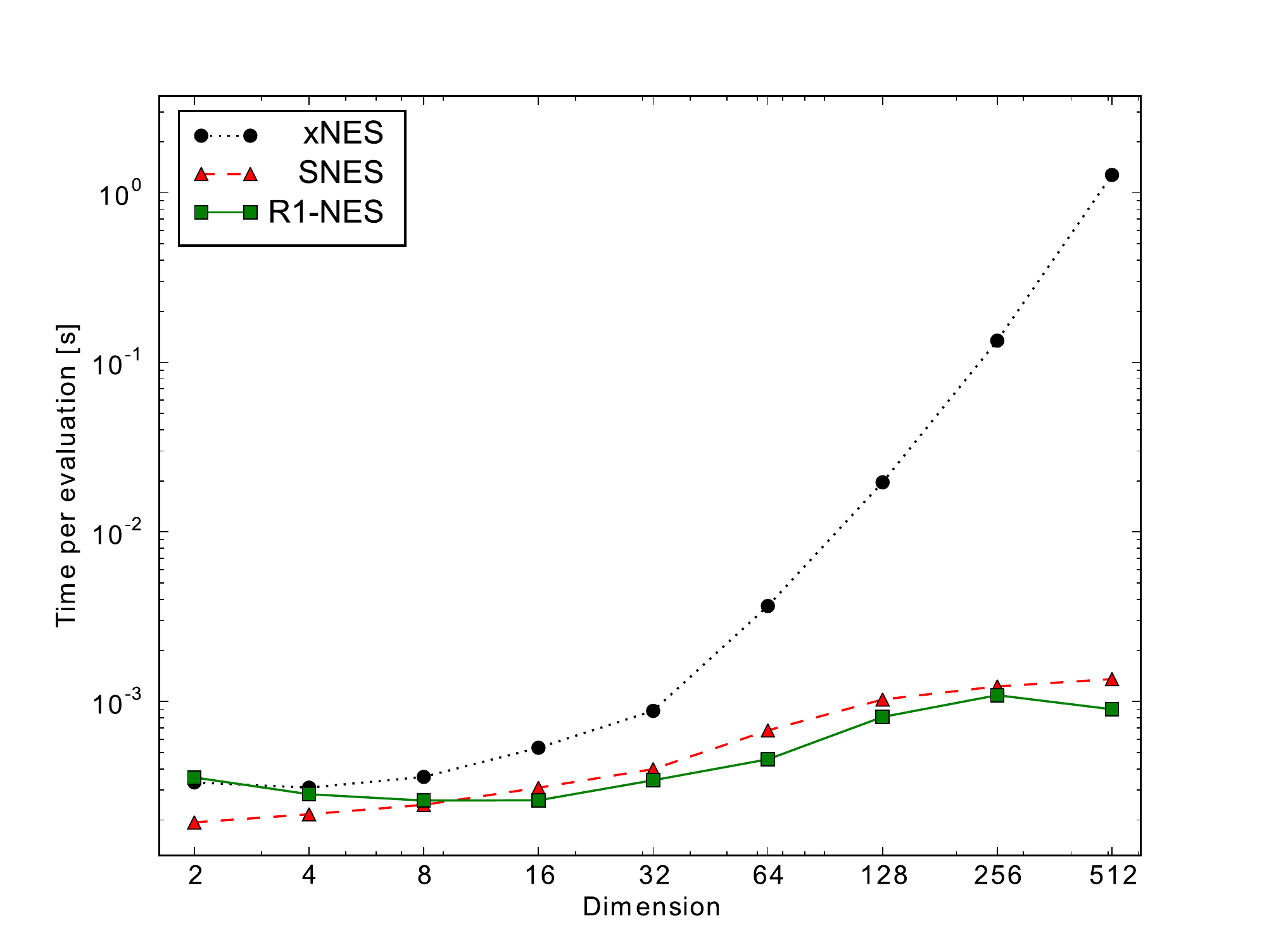}
	} \caption{\textbf{Computation time per function evaluation}, for the three
algorithms, on problem dimensions ranging from 2 to 512. Both SNES and R1-NES
scale linearly, whereas the cost grows cubically for xNES.}%
\label{fig:runtime}%
\end{figure}

\begin{figure}[t]
\begin{center}
\includegraphics[
width=\columnwidth
]{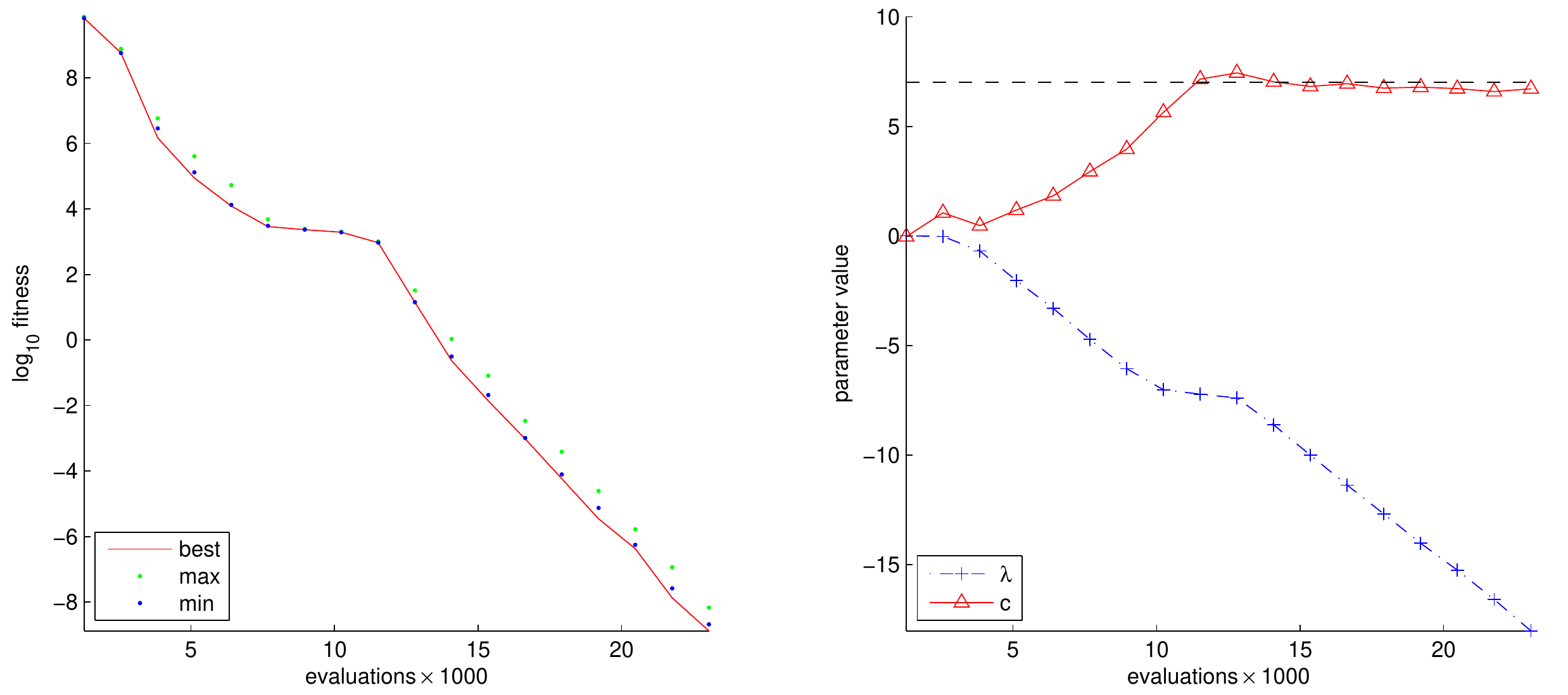}
\end{center}
\caption{\textbf{Behavior of R1-NES on the 32-dimensional cigar function}:
$f\left(  x_{1},\dots,x_{d}\right)  =10^{6}x_{1}^{2}+x_{2} ^{2}+\cdots
+x_{d}^{2}$. The left panel shows the best fitness found so far, and the min
and max fitness in the current population. The right panel shows how $\lambda$
and $c$ evolve over time. Note that the $\lambda$ decreases almost linearly,
indicating that all the other directions except the predominant one shrink
exponentially. In contrast, $c$ first increases, and then stabilizes around
$\log1000$ (the black line). As a result, $I+uu^{\top}$ corresponds to the
Hessian of the cigar function $\left[  10^{6},1,\dots,1\right]  $.}%
\label{fig:cigar}%
\end{figure}

Fig.~\ref{fig:good} shows the results for the eight benchmarks on which R1-NES
performs at least as well as the other methods, and often much better. For
dimensionality under $64$, R1-NES is comparable to xNES in terms of the number
of fitness evaluations, indicating that the rank-one parameterization of the
search distribution effectively captures the local curvature of the fitness
function (see Fig.\ref{fig:cigar} for example). However, the time required to
compute the update for the two algorithms differs drastically, as depicted in
Fig.~\ref{fig:runtime}. For example, a typical run of xNES in $64$ dimensions
takes hours (hence the truncated xNES curves in all graphs), compared to
minutes for R1-NES. As a result, R1-NES can solve these problems up to $512$
dimensions in acceptable time. In particular, the result on the $512$%
-dimensional Rosenbrock function is, to our knowledge, the best to date. We
estimate that optimizing the 512-dimensional sphere function with xNES (or any
other full parameterization method, e.g.\ CMA-ES) would take over a year in
computation time on the same reference hardware. It is also worth pointing out
that sNES, though sharing similar, low complexity per function evaluation, can
only solve separable problems (Sphere, Linear, AttractiveSector, and Ellipsoid).

\begin{figure}[t]
\centerline{
		\includegraphics[width=\columnwidth]{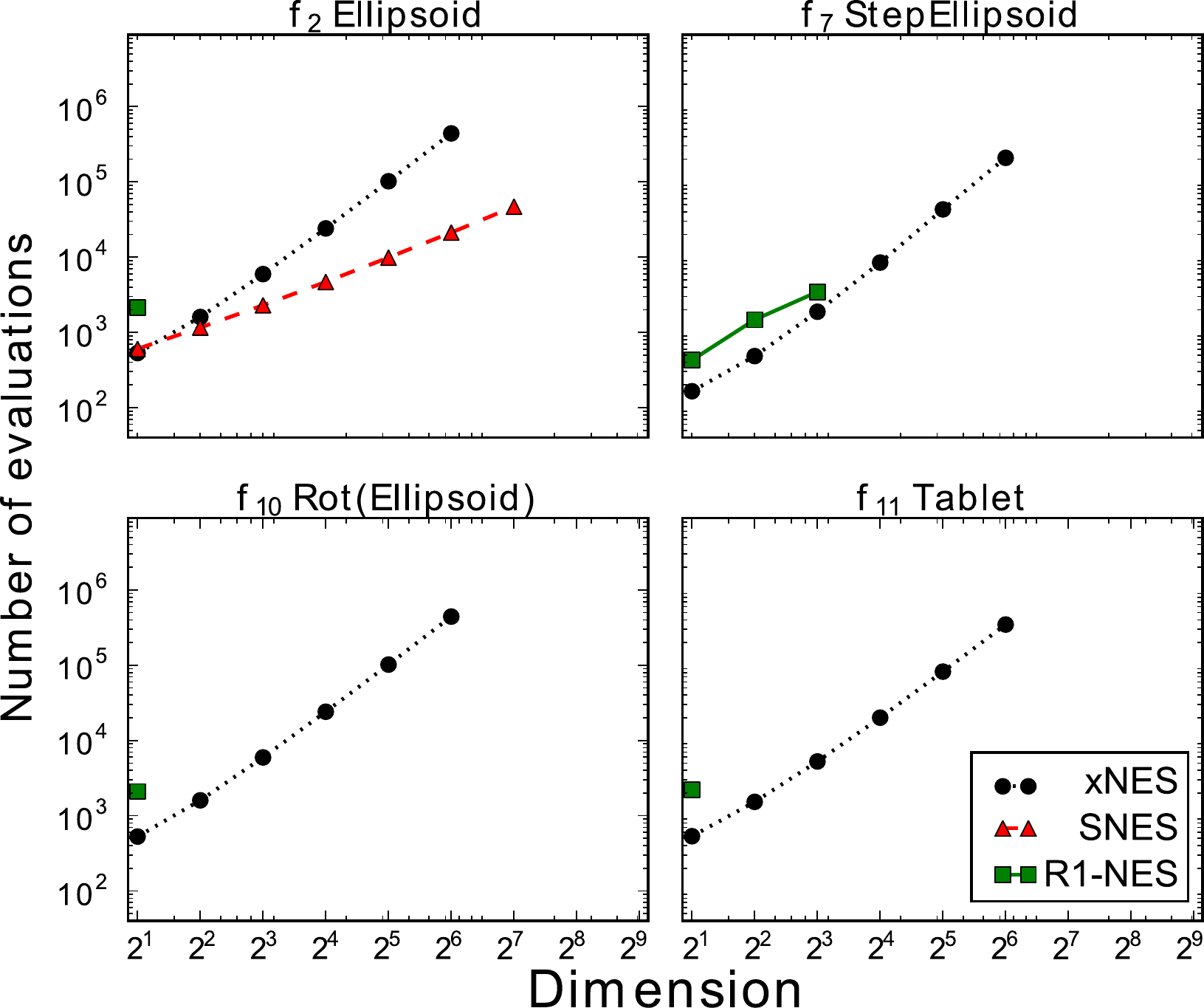}
	} \caption{\textbf{Performance comparison on BBOB unimodal benchmarks for
which R1-NES is not well suited.} For these four functions a single
eigendirection is not enough. Not that SNES solves the Ellipsoid function
because it is separable.}%
\label{fig:bad}%
\end{figure}

Fig.~\ref{fig:bad} shows four cases (Ellipsoid, StepEllipsoid,
RotatedEllipsoid, and Tablet) for which R1-NES is not suited, highlighting a
limitation of the algorithm. Three of the four functions are from the
Ellipsoid family, where the fitness functions are variants of the type%
\[
f\left(  x_{1},\dots,x_{d}\right)  =\sum_{i=1}^{d}x_{i}^{2000\cdot\frac
{i-1}{d}}\text{.}%
\]
The eigenvalues of the Hessian span several orders of magnitude, and the
parameterization with a single predominant direction is not enough to
approximate the Hessian, resulting in poor performance. The other function
where R1-NES fails is the Tablet function where all but a one eigendirection
has a large eigenvalue. Since the parameterization of R1-NES only allows a
single direction to have a large eigenvalue, the shape of the Hessian cannot
be effectively approximated.


\section{Conclusion and future work}

\label{sec:conclusion}

We presented a new black-box optimization algorithm R1-NES that employs a
novel parameterization of the search distribution covariance matrix which
allows a predominant search direction to be adjusted using the natural
gradient with complexity linear in the dimensionality. The algorithm shows
excellent performance in a number of high-dimensional non-separable problems
that, to date, have not been solved with other parameterizations of similar complexity.

Future work will concentrate on overcoming the limitations of the algorithm
(shown in Fig~\ref{fig:bad}). In particular, we intend to extend the algorithm
to a) incorporate multiple search directions, and b) enable each search
direction to \emph{shrink} as well as grow.

\section*{Acknowledgement}
This research was funded in part by Swiss National Science Foundation grants 200020-122124, 200020-125038, and EU IM-CLeVeR project (\#231722).

\bibliographystyle{abbrv}
\bibliography{jmlrbib,self,short}

\end{document}